\documentclass[12pt]{article}
\usepackage[utf8]{inputenc}
\usepackage{amsmath,amsfonts,amssymb,amsthm,epsfig,epstopdf,titling,url,array}
\usepackage{setspace}
\usepackage{multirow}
\usepackage[round]{natbib}
\usepackage[margin=0.9in]{geometry}
\usepackage{xcolor}
\usepackage{authblk}
\usepackage{hyperref}
\usepackage{booktabs}

\usepackage{subcaption}
\usepackage{multirow}
\usepackage{ctable}
\usepackage{array}
\usepackage{colortbl}
\usepackage{multirow}
\usepackage{array}
\usepackage{colortbl}
\usepackage{here}
\usepackage{float}
\usepackage{tabularx}
\usepackage[normalem]{ulem}

\def\bQ{\bfm Q}
\def\cQ{{\cal Q}}

\DeclareMathOperator{\E}{\mathbb{E}}
\newcommand{\bTheta}{\boldsymbol{\Theta}} 

\newcommand{\tr}{\text{tr}}

\newcommand{\Cov}{\text{Cov}}

\newcommand{\bfm}[1]{\ensuremath{\mathbf{#1}}}
\def\ba{\bfm a}

     \def\bD{\bfm D}

\def\bh{\bfm h}          
          
     \def\bJ{\bfm J}

     \def\bQ_\lambda{\bfm Q_\lambda}     \def\cQ_\lambda{{\cal  Q_\lambda}}

     \def\bX{\boldsymbol{X}}

\newcommand{\bfsym}[1]{\ensuremath{\boldsymbol{#1}}}

\def\btheta    {\bfsym \theta}

\def\bpi       {\bfsym \pi}

\def\bpsi       {\bfsym \psi}

\def\bGamma  {\bfsym \Gamma}
\def\bDelta  {\bfsym {\Delta}}
\def\bSigma  {\bfsym \Sigma}
\def\bOmega  {\bfsym \Omega}

\renewcommand{\hat}{\widehat}

\def\Var{ {\text{Var}} }
\def\E{ {\mathbb{E}} }

\newcommand{\vect}[1]{\boldsymbol{#1}} 
\newcommand{\mat}[1]{\mathbf{#1}}      
\newcommand{\rv}[1]{\boldsymbol{#1}}   

\date{}

\title{Moment-Based Anomaly Detection in High-Dimensional Mixed Data}

\begin{document}

\author[1]{Davide Ferrari}
\author[1,2]{Andrea Nicolodi}
\affil[1]{Free University of Bozen--Bolzano, Italy}
\affil[2]{University of Trento, Italy}

	\maketitle

\begin{abstract}
We propose an anomaly detection method for high-dimensional mixed-type data that maps heterogeneous features to a common scale via standardized moment-based transformations, yielding interpretable feature-level anomaly scores. These scores are combined through a dependence-aware aggregation step to construct the Mixed-data Influence (MDI) score, accounting for correlations across features and providing a unified measure of how strongly an observation departs from the reference data distribution. By incorporating higher-order moments, the MDI can detect anomalies that arise not only from location shifts, but also from changes in variability, skewness, and other distributional characteristics. The aggregation mechanism allows our method to capture both sparse anomalies concentrated in a small number of features and weak, coordinated deviations distributed across many features. We empirically investigate the distribution and thresholding behavior of the MDI score across different sample sizes, dimensions, dependence structures, and data types.  Simulation experiments and two real-data applications, to multi-omics and ESG data, evaluate finite-sample calibration, detection performance, and feature-level interpretability.
\end{abstract}

Key words: Mixed-data, outlier detection, moment equations, high-dimensional data, sparse precision matrix.

\section{Introduction}

Anomaly detection aims to identify observations that deviate substantially from the patterns exhibited by a reference population. It is a fundamental problem in statistics and machine learning, with applications ranging from network intrusion detection and financial fraud to industrial monitoring and biomedical data analysis \citep{rousseeuw2003robust,maronna2019robust,zimek2018there,souiden2022survey,olteanu2023meta,li2025survey}. The increasing scale, dimensionality, and heterogeneity of modern datasets have created a growing need for methods that can detect anomalies in complex tabular data while remaining computationally scalable and interpretable.

In statistics, multivariate outlier detection has traditionally been formulated in terms of departures from a reference distribution. Classical robust methods characterize outlyingness through robust location and scatter estimates, including the Minimum Volume Ellipsoid and Minimum Covariance Determinant \citep{rousseeuw1985multivariate,rousseeuw1999fast,cerioli2010multivariate}, while projection-based and regularized approaches have extended this framework to increasingly challenging settings \citep{filzmoser2008outlier,ro2015outlier,oguamalam2024minimum}. A complementary perspective views outlyingness through departures from statistical moments or, more generally, from the estimating equations that characterize the reference population. Classical regression diagnostics quantify the influence of individual observations on score or estimating equations \citep[Ch.~2]{cook1982residuals}, with extensions to generalized linear models \citep{williams1987generalized} and generalized estimating equations \citep{preisser1996deletion,preisser2008note}. Related gradient-based procedures are likewise constructed from score functions evaluated under a reference model \citep{terrell2002gradient,osorio2024gradient}. These developments motivate viewing an anomaly not only as a point that is geometrically distant from the bulk of the data, but more generally as an observation that perturbs the statistical structure of the reference population.

High dimensionality creates additional challenges for this classical perspective. As the number of variables increases, estimating multivariate location, covariance, and higher-order dependence structures becomes increasingly difficult. At the same time, high dimensionality can provide additional information for detection: many individually weak departures may combine to produce a detectable signal. Several methods exploit this phenomenon through scalable tests based on first-moment departures \citep{srivastava2008test,chen2010two}, while others demonstrate the additional information available from higher-order moments \citep{westfall2014kurtosis,pena2001multivariate,pena2007combining}. Thus, high-dimensional anomaly detection requires methods that can aggregate information across variables without allowing dimensionality and dependence to overwhelm the underlying signal.

Machine learning has developed a complementary set of approaches to anomaly detection. Distance- and neighborhood-based methods identify observations that are separated from their local surroundings, with Local Outlier Factor providing a prominent density-based example \citep{breunig2000lof}. Isolation-based methods instead exploit the relative ease with which anomalous observations can be separated through recursive partitioning, as in Isolation Forest \citep{liu2008isolation,liu2012isolation}. One-class classification methods learn a representation or decision boundary for the reference population, including kernel-based approaches and their deep-learning extensions such as Deep SVDD \citep{ruff2018deep}. Representation-learning methods provide a further direction, with autoencoder-based approaches using reconstruction error or latent-space structure to quantify anomalousness \citep{zong2018deep}. More recent methods incorporate auxiliary or self-supervised learning objectives to learn representations specifically suited to anomaly detection \citep{bergman2020classification}. These approaches can capture complex nonlinear structure and are particularly relevant in high-dimensional settings, but their anomaly scores are generally defined through geometric separation, density, reconstruction, or latent representations rather than through interpretable feature-level distributional departures.

A further challenge arises when high-dimensional data are heterogeneous. Modern tabular datasets commonly combine continuous measurements, counts, binary indicators, ordinal variables, and categorical attributes. Such variables differ in their scales and distributional properties and therefore cannot be compared directly through conventional geometric notions of distance. A common practical strategy is to numerically encode categorical variables and then apply a general-purpose anomaly detector. Although convenient, this can make anomaly scores dependent on the chosen encoding and may obscure the distinct statistical meaning of different feature types.

Methods developed specifically for mixed data include correspondence-based representations \citep{wilkinson2017visualizing,greenacre2017correspondence,riani2022}, latent-variable and mixture models \citep{everitt1988analysis,davidow2022factor,marbac2017model,foss2016semiparametric}, and distance- or dissimilarity-based approaches such as Gower-type measures \citep{gower1971general,chavent2014new}. These methods provide principled mechanisms for handling heterogeneous variables, but they are typically developed for low- or moderate-dimensional settings. Conversely, methods designed for high-dimensional anomaly detection generally focus on numerical data and require preprocessing or embedding of categorical variables. Consequently, simultaneously accommodating feature heterogeneity, high dimensionality, dependence, and multiple forms of anomalous behavior remains challenging.

Motivated by these considerations, we introduce the \emph{Mixed-data Influence} (MDI) framework for anomaly detection in high-dimensional heterogeneous data. MDI adopts an estimating-equation perspective in which an anomalous observation is characterized by its contribution to departures from the reference moment structure. Rather than mapping all variables into a common latent representation, MDI constructs feature-specific representations adapted to the statistical type of each variable. Continuous and count features are represented through standardized moment equations, while categorical features are represented through standardized probability components. These transformations place heterogeneous features on a common scale while retaining information about the particular distributional characteristic through which an observation is anomalous.

The resulting feature-level influence measures capture a broader class of departures than location shifts alone, including changes in variability, asymmetry, tail behavior, and categorical probabilities. MDI then aggregates these contributions through a dependence-aware covariance-whitening step to obtain a global anomaly statistic. This aggregation accounts for redundancy among correlated features while allowing weak but coordinated departures across many variables to accumulate into a detectable signal. The resulting statistic is therefore suited to both sparse anomalies, driven by pronounced deviations in a small number of features, and distributed anomalies, characterized by weaker but coherent departures across many dimensions.

An important feature of MDI is that its global score retains an interpretable decomposition into feature-level contributions. The method therefore provides not only an overall measure of anomalousness but also information about which variables and distributional characteristics contribute to a detection. This distinguishes the proposed framework from approaches whose anomaly scores are primarily based on distance, isolation, reconstruction error, or learned latent representations. At the same time, the moment-based construction avoids requiring all feature types to share a common parametric model or measurement scale.

We evaluate MDI across a range of data-generating mechanisms, dimensional regimes, dependence structures, and anomaly types. We investigate both detection performance and the finite-sample behavior of the MDI statistic, including the accuracy of its Gaussian reference thresholds across varying sample sizes and dimensions.  We further illustrate the methodology in two mixed-type empirical settings. The first application considers multi-omics data, characterized by high dimensionality, heterogeneous measurements, substantial cross-feature dependence, and a need to identify the variables driving anomalous observations. The second considers firm-level ESG data to characterize the KPI configurations through which controversy-exposed financial institutions depart from a controversy-free reference population. In both applications, the additive decomposition of MDI provides a direct interpretation of the feature-level contributions underlying each anomaly score.

The remainder of the paper is organized as follows. Section~\ref{sec:methodology} introduces the mixed-data moment representation and the MDI anomaly score. Section~\ref{sec:properties} investigates its empirical distribution and thresholding behavior. Section~\ref{sec:mc} presents Monte Carlo simulation experiments. Finally, Section~\ref{sec:real_data} illustrates the methodology through two empirical applications: multi-omics data in Section~\ref{sec:omics} and Environmental, Social, and Governance (ESG) data in Section~\ref{sec:esg_example}.

\section{Methodology}\label{sec:methodology}

\subsection{Moment-Based Representation and Featurewise Influence}
\label{sec:representation}

Let $\rv{X}^{(1)},\ldots,\rv{X}^{(n)}$ be an i.i.d. reference sample from the random vector $\rv{X}=(X_1,\ldots,X_d)^\top$, where each feature $X_j$ may be continuous, discrete, or categorical, with support $\mathcal{X}_j$. We represent feature $j$ through a feature-specific map $\bpsi_j:\mathcal{X}_j\to\mathbb{R}^{p_j}$ and its corresponding reference moment
\begin{equation}
\label{eq:moment}
\btheta_{0j}:=\E\!\left[\bpsi_j(X_j)\right]\in\mathbb{R}^{p_j}.
\end{equation}
The map $\bpsi_j$ specifies the distributional characteristics used to assess anomalousness, including location, scale, higher-order moments, or category probabilities. Stacking these representations gives
$\btheta_0=(\btheta_{01}^\top,\ldots,\btheta_{0d}^\top)^\top\in\mathbb{R}^p$,
where $p=\sum_{j=1}^d p_j$.

The feature-specific construction accommodates heterogeneous data types without imposing a common parametric representation. For numerical features, we may take $\bpsi_j(x_j)=(x_j,x_j^2,\ldots,x_j^{k_j})^\top$,
so that the representation incorporates moments beyond location. For a categorical feature with $\ell_j$ levels, we instead define $\bpsi_j(x)=
\big(\mathbf{1}\{x=1\},\ldots,\mathbf{1}\{x=\ell_j-1\}\big)^\top$, giving
$\btheta_{0j}=(\pi_{j1},\ldots,\pi_{j,\ell_j-1})^\top$, $\pi_{jr}=\Pr(X_j=r)$.
Thus, categorical variables are represented through their probability structure rather than through an arbitrary numerical encoding.

Define the featurewise moment departure of observation $i$ by
\begin{equation}
\label{eq:standardized-score}
\rv{S}_j^{(i)}
=
\bpsi_j(X_j^{(i)})-\btheta_{0j},
\qquad
\rv{Z}_j^{(i)}
=
\mathbf{J}_j^{-1/2}\rv{S}_j^{(i)},
\end{equation}
where $\mathbf{J}_j:=\Var\!\left(\bpsi_j(X_j)\right)$. We assume that $\mathbf{J}_j$ is positive definite. The transformation by $\bJ_j^{-1/2}$ removes the scale and dependence
among the moment components used to represent feature $j$.

The statistical interpretation of $\rv{S}_j^{(i)}$ follows directly from the influence function of the moment functional
$
T_j(F)=\E_F[\bpsi_j(X_j)]$.
Indeed, under the contaminated distribution
$F_\varepsilon=(1-\varepsilon)F+\varepsilon\delta_x$,
\[
\left.
\frac{d}{d\varepsilon}T_j(F_\varepsilon)
\right|_{\varepsilon=0}
=
\bpsi_j(x)-\E_F[\bpsi_j(X_j)].
\]
Hence, $\rv{S}_j^{(i)}$ is the influence function of the feature-specific moment functional, while $\rv{Z}_j^{(i)}$ is its standardized version.

We summarize the magnitude of this standardized moment departure by
\begin{equation}
\label{eq:quad2}
Q_j^{(i)}
:=
\|\rv Z_j^{(i)}\|_2^2-p_j
=
\rv S_j^{(i)\top}
\bJ_j^{-1}
\rv S_j^{(i)}
-p_j.
\end{equation}
Under the reference distribution, $\E[Q_j^{(i)}]=0$.
Thus, $Q_j^{(i)}$ is a centered quadratic measure of departure in the
moment space associated with feature $j$. Large positive values indicate that observation $i$ exhibits an unusually large departure from the reference moment profile of feature $j$. Because multiple moments may enter $\bpsi_j$, the measure can respond to departures in location, variability, asymmetry, tail behavior, or category probabilities, depending on the chosen representation.

This interpretation can be made explicit by considering how the expected feature-level score changes when the candidate distribution departs from the reference population. Suppose that, for feature $j$, $\E[\bpsi_j(X_j)]=\btheta_{0j}+\bDelta_j$ and $\Var(\bpsi_j(X_j))=\bJ_j^\star$, while $\bJ_j$ denotes the corresponding reference covariance. Then $\E[Q_j^\star]
=
\tr\!\bigl(\bJ_j^{-1}\bJ_j^\star\bigr)-p_j
+
\bDelta_j^\top\bJ_j^{-1}\bDelta_j$.
Hence, the expected feature-level influence can increase through either a change in the variability of the moment representation or a displacement of its mean. Depending on the choice of $\bpsi_j$, these departures may correspond to changes in location, scale, skewness, higher-order moments, or categorical probabilities.

Collecting the feature-level scores gives
$\rv Q^{(i)}
=
\left(
Q_1^{(i)},\ldots,Q_d^{(i)}
\right)^\top
\in\mathbb R^d$.
Although the components of $\rv Q^{(i)}$ are centered and each is based
on a within-feature standardized moment representation, they need not
have equal variances and may remain dependent across features.

In practice, $\btheta_{0j}$ and $\mathbf{J}_j$ are unknown and are estimated from the reference sample:
\[
\hat{\btheta}_{j}
=
\frac{1}{n}\sum_{r=1}^n\bpsi_j(X_j^{(r)}),
\qquad
\hat{\mathbf{J}}_{j}
=
\frac{1}{n}\sum_{r=1}^n
\big(\bpsi_j(X_j^{(r)})-\hat{\btheta}_{j}\big)
\big(\bpsi_j(X_j^{(r)})-\hat{\btheta}_{j}\big)^\top.
\]
For a candidate observation $\rv{X}^\star$, define
$
\hat{\rv{S}}_j^\star
=
\bpsi_j(X_j^\star)-\hat{\btheta}_j$ and
$\hat{\rv{Z}}_j^\star
=
\hat{\mathbf{J}}_j^{-1/2}\hat{\rv{S}}_j^\star$,
$\hat{Q}_j^\star
=
\|\hat{\rv{Z}}_j^\star\|_2^2-p_j$.
The reference sample is therefore used exclusively to estimate the baseline representation, while the candidate observation is evaluated relative to that independently estimated reference distribution. Collecting the featurewise contributions gives
$\hat{\vect{Q}}^\star
=
(\hat{Q}_1^\star,\ldots,\hat{Q}_d^\star)^\top
\in\mathbb{R}^d$.
Although each $\hat{Q}_j^\star$ is constructed marginally, the vector $\hat{\vect{Q}}^\star$ retains the dependence induced by the joint distribution of the features. The subsequent aggregation step exploits this dependence to combine featurewise evidence into a global anomaly statistic.

\subsection{Dependence-Aware Aggregation}
\label{sec:aggregation}

The feature-level representation $\rv Q=(Q_1,\ldots,Q_d)^\top$ provides a collection of scalar measures of departure from the reference moment structure. The next step is to combine these contributions into a global anomaly score. Simply summing the components would implicitly treat the information carried by different features as non-redundant. This can be undesirable in high-dimensional settings, where correlated features may exhibit similar influence patterns and thereby repeatedly reflect the same underlying departure. Conversely, an anomalous observation may exhibit many individually moderate deviations whose joint behavior is substantially more unusual than any single feature-level contribution. We therefore construct an aggregation rule that explicitly accounts for dependence among the components of $\rv Q$.

Let $\bSigma:=\Cov(\rv Q)$
denote the covariance matrix of the feature-level influence vector under the reference distribution, and define
$
\bD_
:=
\operatorname{diag}
\bigl(
(\bSigma)_{11},\ldots,(\bSigma)_{dd}
\bigr)$. We standardize the marginal variability of the feature-level scores by
setting $\rv R
:=
\bD^{-1/2}\rv Q$.
Since $\E(\rv Q)=\mathbf 0_d$, we also have
$\E(\rv R)=\mathbf 0_d$, while
\[
\bGamma
:=
\Cov(\rv R)
=
\bD^{-1/2}\bSigma\bD^{-1/2}
=
\text{Corr}(\rv Q).
\]
with corresponding precision matrix by
$\bOmega
:=
\bGamma^{-1}$. The matrices $\bJ_j$ and $\bD$ therefore play distinct roles.
The matrix $\bJ_j$ standardizes the moment components within feature
$j$, whereas $\bD$ standardizes the marginal variability of the
resulting scalar feature-level scores $Q_j$.  

We use the standardized representation $\rv R$ to learn the dependence
structure among the feature-level scores. Taking $\mathbf 1_d$ as the
benchmark direction for a common positive departure, define
\[
\ba^\star
=
\arg\max_{\ba\in\mathbb R^d} \
\ba^\top\mathbf 1_d
\qquad
\text{subject to}
\qquad
\ba^\top\bGamma\ba=1.
\]
The corresponding solution is
\[
\ba^\star
=
\frac{\bOmega\mathbf 1_d}
{\sqrt{\mathbf 1_d^\top\bOmega\mathbf 1_d}}.
\]
The dependence-adjusted weights $\ba^\star$ are then applied to the
original feature-level influence scores,
$U
=
(\ba^\star)^\top\rv Q$.
Under the reference distribution, $\E(U)=0$,
$\Var(U)
=
(\ba^\star)^\top\bSigma\ba^\star$.
Thus, the correlation-standardized representation is used to determine
the aggregation weights, while the final projection is formed on the
original feature-level score scale.

In practice, the required quantities are unknown and must be estimated from the reference sample. To reduce the effect of evaluating a reference observation using moment estimates to which that same observation contributed, we construct leave-one-out feature-level influence vectors. For $r=1,\ldots,n$, let
$
\hat{\rv Q}_{r}^{(-r)}
=
(\
\hat Q_{1,r}^{(-r)},\ldots,
\hat Q_{d,r}^{(-r)}
)^\top
$
denote the feature-level influence representation of reference observation $r$, computed using feature-specific moment and covariance estimates obtained from the reference sample with observation $r$ removed. Let $\hat\bD$ denote the diagonal matrix of empirical
marginal variances of the leave-one-out feature-level score vectors
$\hat{\rv Q}_r^{(-r)}$, and define
$\hat{\rv R}_r^{(-r)}
=
\hat\bD^{-1/2}\hat{\rv Q}_r^{(-r)}$
The correlation matrix $\bGamma$ is estimated from
$\hat{\rv R}_1^{(-1)},\ldots,\hat{\rv R}_n^{(-n)}$, yielding the
precision-matrix estimate $\hat\bOmega$. We then define the empirical directions
\[
\hat\ba
=
\frac{\hat\bOmega\mathbf 1_d}
{\sqrt{\mathbf 1_d^\top\hat\bOmega\mathbf 1_d}}.
\]

For a candidate observation $\rv X^\star$, the fitted aggregate
influence is
$
\hat U^\star
=
\hat\ba^\top\hat{\rv Q}^\star$.
Since the variance normalization used to construct $\hat\ba$ pertains
to the standardized representation $\hat{\rv R}$ rather than directly
to $\hat{\rv Q}$, the variability of the fitted projection is calibrated
from its leave-one-out reference distribution. 

The Mixed-data Influence score of $\rv{\bX}^\star$ is then defined as
\[
\operatorname{MDI}(\rv X^\star)
=
\frac{
\hat\ba^\top\hat{\rv Q}^\star
}{
\hat\sigma_{\mathrm{LOO}}
},
\]
where $\hat\sigma_{\mathrm{LOO}}$ denotes the sample standard deviation
of $W_1,\ldots,W_n$, with 
$W_r
=
\hat\ba^\top\hat{\rv Q}_r^{(-r)}$, $r=1,\ldots,n$. The MDI score summarizes the extent to which a candidate observation departs from the reference moment structure after accounting for dependence among the feature-level scores. Large positive values indicate departures that align strongly with the dependence-adjusted aggregation direction. Because the numerator is a linear combination of the feature-level scores, the statistic retains an additive decomposition, so that both the overall anomaly score and the individual contributions $\hat Q_j^\star$ can be examined. Since each $\hat Q_j^\star$ is derived from a feature-specific moment representation, these contributions also indicate which variables and distributional characteristics drive the detected departure.

 The construction separates dependence adjustment at two distinct levels.
For feature $j$, the matrix $\mathbf J_j\in\mathbb R^{p_j\times p_j}$
accounts for dependence among the moment components used to construct
$Q_j$, whereas $\bGamma\in\mathbb R^{d\times d}$ describes dependence
across the standardized feature-level scores. This avoids estimating a
single covariance matrix over all $p=\sum_{j=1}^d p_j$ moment components,
which may be difficult in heterogeneous high-dimensional settings. At
the same time, the feature-level representation remains available for
interpretation, and the linear aggregation can accommodate both
concentrated anomalies, driven by a small number of large feature-level
departures, and distributed anomalies, in which weaker but coordinated
departures accumulate across many features.

The normalization by $\hat\sigma_{\mathrm{LOO}}$ standardizes the fitted projection with respect to its leave-one-out reference variability. The role of the two standardizations is distinct: the correlation-standardized representation $\hat{\rv R}$ is used to determine the relative aggregation weights through $\hat\bOmega$, whereas these weights are subsequently applied to the original feature-level scores $\hat{\rv Q}$. Consequently, the variance normalization imposed when constructing $\hat\ba$ on the $\hat{\rv R}$ scale is not preserved after projecting $\hat{\rv Q}$, and the resulting aggregate need not have unit variance. Let $\hat\bSigma_{\mathrm{LOO}}$ denote the empirical covariance matrix of $\hat{\rv Q}_1^{(-1)},\ldots,\hat{\rv Q}_n^{(-n)}$. Since $W_r=\hat\ba^\top\hat{\rv Q}_r^{(-r)}$, its leave-one-out variance is
$\hat\sigma_{\mathrm{LOO}}^2
=
\hat\ba^\top\hat\bSigma_{\mathrm{LOO}}\hat\ba$.
Thus, the final scaling simply normalizes the fitted aggregation direction on the original $\hat{\rv Q}$ scale, so that the resulting projection has unit empirical reference variance. When $\hat\bOmega$ is obtained by direct inversion of the empirical correlation matrix, the relation $\hat\bOmega=\hat\bD^{1/2}\hat\bSigma_{\mathrm{LOO}}^{-1}\hat\bD^{1/2}$ makes the connection between the correlation- and covariance-scale representations explicit. This identity, however, does not generally hold for regularized precision estimators. The LOO normalization, therefore, separates dependence adjustment from final score calibration and places MDI values on a common reference scale, which is particularly useful when a common detection threshold is employed.

\subsection{Learning the Dependence Structure}
\label{sec:precision}

When $d$ is small relative to $n$, the dependence structure of the
feature-level influence representation can be estimated directly from the
sample correlation matrix of the standardized scores
$\hat{\rv R}_1^{(-1)},\ldots,\hat{\rv R}_n^{(-n)}$, with the corresponding
precision matrix obtained by inversion. However, when $d$ is comparable
to or larger than $n$, the empirical correlation matrix may be poorly
conditioned or singular, making direct inversion unstable or infeasible.
Regularized estimation is also natural in many applications where
dependence is expected to be sparse, in the sense that each feature-level
score is associated with only a limited subset of the remaining scores.
Such a structure is plausible when groups of features are correlated
within a small number of related blocks or local neighborhoods, while
most feature pairs are only weakly related. We therefore use a sparse
precision-matrix estimator to obtain the dependence-adjusted aggregation
direction in these settings.

For concreteness, we estimate $\bOmega=\bGamma^{-1}$ using the graphical lasso of \citet{friedman2008sparse}. Let
$
\hat\bGamma^{\mathrm{emp}}
=
(\hat\gamma_{jk})_{1\leq j,k\leq d}$
denote the empirical correlation matrix computed from the leave-one-out feature-level influence vectors
$\hat{\rv Q}_{1}^{(-1)},\ldots,\hat{\rv Q}_{n}^{(-n)}$.
For a regularization parameter $\lambda>0$, the graphical-lasso estimator is defined by
\[
\hat\bOmega_{\lambda}
=
\arg\min_{\bTheta\succ0}
\left\{
\tr\!\left(\hat\bGamma^{\mathrm{emp}}\bTheta\right)
-\log\det(\bTheta)
+\lambda\sum_{j\neq k}|\Theta_{jk}|
\right\}.
\]
The $\ell_1$ penalty on the off-diagonal entries favors a sparse precision matrix and stabilizes estimation when the dimension is large relative to the reference sample size.

Within the MDI framework, $\hat\bOmega_\lambda$ is used primarily to construct the dependence-adjusted aggregation direction
\[
\hat\ba_\lambda
=
\frac{\hat\bOmega_\lambda\mathbf 1_d}
{\sqrt{\mathbf 1_d^\top
\hat\bOmega_\lambda\mathbf 1_d}}.
\]
Thus, the objective is not necessarily to recover the conditional-independence graph of the original features. Rather, the precision matrix provides a regularized representation of dependence among the feature-level influence measures, which is then used to adjust their aggregation in the global anomaly score.

The regularization parameter $\lambda$ determines the degree of sparsity imposed on the estimated dependence structure. Larger values produce sparser precision estimates by shrinking more off-diagonal entries toward zero, whereas smaller values retain a richer dependence structure at the cost of potentially greater estimation variability. The procedure used to select $\lambda$ is described in Section~\ref{sec:mc}, where we also investigate the sensitivity of the resulting MDI scores to regularization across different dimensional and dependence regimes.

Although we use the graphical lasso throughout, the MDI construction does not depend on this particular estimator. Alternative regularized covariance or precision-matrix estimators can be incorporated without modifying either the feature-level moment representation or the subsequent scoring mechanism.

\subsection{Reference Distribution and Calibration}
\label{sec:properties}

The MDI score is continuous and may be used either to rank candidate
observations or, for binary detection, to flag observations satisfying
$\operatorname{MDI}(\rv X^\star)>c$, where $c$ is chosen according to the
desired operating point. Thresholding is therefore only one possible use
of the score. To motivate a Gaussian reference threshold, let
$\sigma_j^2=\Var(Q_j)$ and $R_j=Q_j/\sigma_j$, so that
$\E(R_j)=0$ and $\Var(R_j)=1$. Since
$\rv Q=\bD^{1/2}\rv R$, the standardized population projection can be
written as
\[
\operatorname{MDI}(\rv X^\star)
=
\sum_{j=1}^d \alpha_j R_j^\star,
\qquad
\alpha_j
=
\frac{a_j^\star\sigma_j}
{\sqrt{(\ba^\star)^\top\bSigma\ba^\star}},
\]
and has mean zero and variance one under the reference distribution.
Thus, as $d$ increases, MDI is a standardized weighted sum of
feature-level contributions. Under sufficiently controlled moments, weak
enough residual dependence, and the absence of dominant coordinates,
central-limit arguments for weighted dependent sums motivate the
approximation $\operatorname{MDI}(\rv X^\star)\approx\mathcal N(0,1)$.
Standardization alone does not imply Gaussianity or independence, however,
and the approximation may deteriorate in the presence of strong
higher-order dependence, heavy-tailed contributions, or highly
concentrated weights.

For binary detection, this motivates the reference threshold $c_\alpha=z_{1-\alpha}$,
where $z_{1-\alpha}$ is the $(1-\alpha)$ quantile of the standard normal
distribution. When the approximation is accurate, the rule
$\operatorname{MDI}(\rv X^\star)>z_{1-\alpha}$ has reference
false-positive probability approximately $\alpha$. Other thresholds may
instead be obtained from empirical reference scores or chosen according
to application-specific operating criteria. In Section~\ref{sec:mc}, we
assess the adequacy of the Gaussian calibration by comparing the empirical
null distribution and upper quantiles of MDI with their standard normal
counterparts and by reporting the realized false-positive rates associated
with $z_{1-\alpha}$ across different dimensions, marginal distributions,
and dependence structures.

\section{Monte Carlo Experiments} \label{sec:mc}

We assess finite-sample performance under varying proportions of continuous, discrete, and categorical features. Feature-level distributions span four representative families:
\begin{itemize}
\item[(A)] \textit{Normal:} $X_j\sim \mathcal N(\mu_0, \tau_0^2)$, $(\mu_0, \tau_0^2)=(0,1)$,
$\bpsi_j(x)=(x, x^2)^\top $,  $p_j=2$;
\item[(B)] \textit{Skew–Normal:} $X_j \sim \mathrm{SN}(\xi_0,\omega_0,\alpha_0)$ with
$(\xi_0,\omega_0,\alpha_0)=(0,1,5)$ (positively skewed), with
$\bpsi_j(x)=(x,\ x^2,\ x^3)^\top$, so $p_j=3$.
\item[(C)] \textit{Poisson:} $X_j\sim \mathrm{Pois}(\lambda_0)$ with $\lambda_0=4$, $\psi_j(x)=x$, $p_j=1$;
\item[(D)]  \textit{Categorical with 3 levels:} $X_j\in\{1,2,3\}$ with baseline probabilities
$\bpi_0 = (\pi_{01},\pi_{02},\pi_{03})=(0.5,0.3,0.2)$,
$\bpsi_j(x)= (\vect 1\{x=1\}, \vect 1\{x=2\})^\top$, $p_j=2$.
\end{itemize}
Feature dependence is induced through a Gaussian copula. Specifically, we generate
$\rv V^{(i)}=(V_1^{(i)},\dots,V_d^{(i)})^\top \sim \mathcal N_d(\vect 0,\mat C)$, set
$U_j^{(i)}=\Phi(V_j^{(i)})$ componentwise, where \(\Phi(\cdot)\) denotes the standard normal cdf, and then define $
X_j^{(i)} = F_j^{\leftarrow}\!\bigl(U_j^{(i)}\bigr)$,
where \(F_j^{\leftarrow}\) denotes the generalized inverse cdf, or quantile map, associated with the chosen marginal distribution of feature \(j\). For categorical variables, this is understood with respect to a fixed ordering of the category labels. 

To examine the role of cross-coordinate dependence, we consider two models for $\mat{C}$, each representing a different sparsity regime:  (i) \textit{Blockwise constant (BC) structure.}
$\mat{C}$ is block diagonal with $B=d/b$ equal-sized blocks of dimension $b=5$, $
\mat{C}_b = (1-\rho_0)\,\mat I_b + \rho_0\,\mathbf 1_b\mathbf 1_b^\top$, $\rho_0=0.5$. (ii) \textit{Autoregressive (AR) structure.}
$\mat{C}$ has entries $\mat C_{jk}=\rho_0^{|j-k|}$ with $\rho_0=0.5$. The BC model corresponds to a strictly sparse structure, where each coordinate interacts with only a fixed number of others. The AR structure is not strictly sparse, but its correlations decay geometrically, yielding an effective row sparsity comparable to that of the block-constant model.

In each Monte Carlo replication, the training data are first generated under the reference model described above, with sample size \(n=150\) and dimension $d=20,30, 200, 300$. 
We then generate one independent candidate observation from the same Gaussian-copula model. Under the null, the candidate observation is left unchanged. Under the alternative, contamination is introduced directly in its moment representation, thereby providing a common perturbation mechanism across the different data types.

For the candidate observation, contamination is introduced through feature-specific moment shifts $\btheta_j^{(i)}=\btheta_{0j}+\bDelta_j^{(i)}$, $
\bDelta_j^{(i)}=\mat J_{0j}^{1/2}\bh_j$,
where $\bh_j\in\mathbb R^{p_j}$ denotes the standardized moment-shift vector for feature $j$. This parametrization expresses departures on the reference-standardized moment scale, so that the magnitude of $\bh_j$ is comparable across heterogeneous feature types.
Under this parametrization, we consider two contamination regimes:

\begin{itemize}
\item \emph{Dense small shifts.} We set $\bh_j = h_0 \,\vect{1}/{d^{\frac{1}{k}}}$ for $k \in \{2,3,4\}$ and all $j \leq d$.
In this case each feature carries a vanishing per-feature signal of order $O(d^{-1/k})$, so that no single coordinate is detectable. However, their cumulative effect aggregates to a nontrivial noncentrality parameter $\delta^{(i)}$, in direct analogy with Pitman local
alternatives.

\item \emph{Sparse fixed-amplitude shifts.} Only $s\ll d$ features are contaminated, each with constant standardized signal $\bh_j = h_0 \vect{1}$. Under bounded moment dimensions and a stable dependence structure, the resulting standardized aggregate displacement is heuristically of order $s h_0^2/\sqrt d$.
\end{itemize}

In all simulation experiments, the standardized precision matrix $\bOmega=\bGamma^{-1}$is estimated using graphical lasso estimator, with the tuning parameter \(\lambda\) selected by minimizing the extended Bayesian information criterion \citep{chen2008extended}, $
\mathrm{EBIC}(\lambda)
=
-2\,\ell(\hat\bOmega_\lambda)
+
|E_\lambda|\log n
+
4\gamma\,|E_\lambda|\log d$, $\gamma\in[0,1]$,
where
$
\ell(\hat\bOmega_\lambda)
=
\frac{n}{2}
\left\{
\log\det(\hat\bOmega_\lambda)
-
\tr(\hat\bGamma^{\mathrm{emp}}\,\hat\bOmega_\lambda)
\right\}$
is the Gaussian log-likelihood, up to an additive constant, and
\(E_\lambda=\{(j,k):\, j<k,\ (\hat\bOmega_\lambda)_{jk}\neq 0\}\) is the estimated edge set.
We set $\gamma=0.5$, so that the final EBIC penalty term reduces to $2|E_\lambda|\log d$. Here, $\hat\bGamma^{\mathrm{emp}}$ is the empirical correlation matrix of the leave-one-out quadratic scores defined in Section~\ref{sec:precision}.
All analyses were conducted in \textsf{R} \citep{RCoreTeam2024}. Graphical-model estimation was performed using the \texttt{huge} package \citep{zhao2012huge}.

\begin{table}[ht]
\centering
\begin{tabular}{l cccc c cccc}
\hline
& \multicolumn{4}{c}{Block correlation} & & \multicolumn{4}{c}{ {AR(1) correlation}} \\
\cmidrule(r){2-5} \cmidrule(l){6-10}
$k$ & $d=20$ & $d=30$ & $d=200$ & $d=300$ & & $d=20$ & $d=30$ & $d=200$ & $d=300$\\
\midrule
\multicolumn{10}{c}{Balanced mix (baseline): A:B:C:D = 25:25:25:25} \\
0 & 0.032 & 0.035 & 0.042  & 0.042 & & 0.027 & 0.034 & 0.054 &  0.055 \\
2 & 0.728 & 0.598 & 0.081 & 0.078 & & 0.748 & 0.565 & 0.076 & 0.058 \\
3 & 0.956 & 0.937 & 0.813 & 0.756 & & 0.955 & 0.951 & 0.820& 0.776 \\
4 & 0.978  & 0.981 & 0.983 & 0.984 & & 0.979 & 0.985 & 0.988 & 0.980 \\

\multicolumn{10}{c}{Numeric-heavy: A:B:C:D = 40:40:10:10} \\
0 & 0.035 & 0.036 & 0.056 & 0.055 & & 0.044 & 0.037 & 0.053 &  0.071 \\
2 & 0.632 & 0.452 & 0.074 & 0.072 & & 0.673 & 0.475 & 0.082 & 0.081  \\
3 & 0.939 & 0.931 & 0.776 & 0.707 & & 0.921 & 0.924 & 0.785 & 0.693 \\
4 & 0.981 & 0.983 & 0.971 & 0.979 & & 0.978 & 0.975 & 0.981 & 0.984\\

\multicolumn{10}{c}{Categorical-heavy: A:B:C:D = 10:10:10:70} \\

0 & 0.026 & 0.030  & 0.034 & 0.051 & & 0.027 & 0.030 & 0.033 & 0.035 \\
2 & 0.916 & 0.844  & 0.183 & 0.095 & & 0.900 & 0.851 & 0.167 & 0.105 \\
3 & 0.983 & 0.978 & 0.925 & 0.935 & & 0.979 & 0.978 & 0.954 & 0.923 \\
4 & 0.993 & 0.997 & 0.993 & 0.995 & & 0.992 & 0.994 & 0.991 & 0.994 \\

\multicolumn{10}{c}{Skew-extreme: A:B:C:D = 10:70:10:10} \\

0 & 0.036 & 0.046 & 0.076 & 0.093 & & 0.040 & 0.044 & 0.078 & 0.094 \\
2 & 0.479 & 0.297 & 0.105 & 0.135 & & 0.468 & 0.334 & 0.090 & 0.115 \\
3 & 0.902 & 0.893 & 0.777 & 0.758 & & 0.912 & 0.882 & 0.768 & 0.771  \\
4 & 0.970 & 0.982 & 0.976 & 0.986 & & 0.971 & 0.978 & 0.987 & 0.981  \\

\midrule
\end{tabular}
\caption{Monte Carlo estimates of Type I error ($k = 0$) and power ($k > \in \{2,3,4\}$) for the global test with size $\alpha = 0.05$ under dense shift contamination regime.  For $k\in \{2,3,4\}$, all features are contaminated according to $\btheta_j^{(i)}=\btheta_{0j}+\bDelta_j^{(i)}$, where $\bDelta_j^{(i)}=\mat J_{0j}^{1/2}\bh_j$ and $\bh_j=10\vect 1/d^{1/k}$, $j\le d$. The row $k=0$ corresponds to the null model, with $\bh_j=\vect 0$ for all $j\le d$. Results are reported for different proportions of variable types: Normal (A), Skew-Normal (B), Poisson (C) and categorical (D) under block and AR(1) correlation structures. Results are based on $1000$ Monte Carlo sample of size $n=150$.}
\label{table1}
\end{table}

\clearpage

\begin{table}[ht]
\centering
\begin{tabular}{l cccc c cccc}
\midrule
& \multicolumn{4}{c}{Block correlation} & & \multicolumn{4}{c}{AR(1) correlation} \\
\cmidrule(r){2-5} \cmidrule(l){7-10}
$s/d$ & $d=20$ & $d=30$ & $d=200$ & $d=300$ & & $d=20$ & $d=30$ & $d=200$ & $d=300$ \\
\midrule

\multicolumn{10}{c}{Balanced mix (baseline): A:B:C:D = 25:25:25:25} \\
0 & 0.035 & 0.034 & 0.045 & 0.040 & & 0.025 & 0.038  & 0.051 & 0.056 \\
0.1 & 0.894 & 0.923 & 0.990 & 0.996 & & 0.907 & 0.919 & 0.992 & 0.998 \\
0.2 & 0.962 & 0.985 & 0.999 & 0.998 & & 0.975 & 0.985 & 0.999 & 1.000 \\
0.3 & 0.995 & 0.995 & 1.000 & 1.000 & & 0.984 & 0.992 & 1.000 & 1.000 \\

\multicolumn{10}{c}{Numeric-heavy: A:B:C:D = 40:40:10:10} \\
0 & 0.033 & 0.038 & 0.058 & 0.051 & & 0.023 & 0.041 & 0.055 & 0.058 \\
0.1 & 0.864 & 0.908 & 0.994 & 0.995 & & 0.885 & 0.921 & 0.993 & 1.000 \\
0.2 & 0.960 & 0.976 & 0.997 & 0.999 & & 0.963 & 0.983 & 0.999 & 1.000 \\
0.3 & 0.982 & 0.993 & 1.000 & 1.000 & & 0.990 & 0.989 & 1.000 & 1.000 \\

\multicolumn{10}{c}{Categorical-heavy: A:B:C:D = 10:10:10:70} \\
0 & 0.029 & 0.030 & 0.030 & 0.039 & & 0.031 & 0.025 & 0.044 & 0.039 \\
0.1 & 0.962 & 0.967 & 0.996 & 0.996 & & 0.966 & 0.966 & 0.997 & 1.000 \\
0.2 & 0.994 & 0.994 & 1.000 & 0.998 & & 0.988 & 0.995 & 1.000 & 1.000 \\
0.3 & 0.994 & 0.995 & 1.000 & 1.000 & & 0.997 & 0.996 & 1.000 & 1.000 \\

\multicolumn{10}{c}{Skew-extreme: A:B:C:D = 10:70:10:10} \\
0 & 0.029 & 0.037 & 0.083 & 0.086 & & 0.035 & 0.035 & 0.084 & 0.095 \\
0.1 & 0.794 & 0.865 & 0.995 & 0.995 & & 0.803 & 0.870 & 0.991 & 0.997 \\
0.2 & 0.953 & 0.976 & 1.000 & 1.000 & & 0.948 & 0.974 & 0.999 & 0.999 \\
0.3 & 0.977 & 0.991 & 1.000 & 1.000 & & 0.990 & 0.991 & 1.000 & 1.000 \\
\midrule
\end{tabular}
\caption{Monte Carlo estimates of Type I error ($s/d=0$) and power ($s/d>0$) for the global test at nominal level $\alpha=0.05$ under the sparse shift contamination regime. For $s>0$, a subset $\mathcal S\subset{1,\ldots,d}$ of $s$ features is selected at random in each replication. The contaminated features satisfy $\btheta_j^{(i)}=\btheta_{0j}+\bDelta_j^{(i)}$, where $\bDelta_j^{(i)}=\mat J_{0j}^{1/2}\bh_j$ and $\bh_j=10,\vect 1$ for $j\in\mathcal S$, while $\bh_j=\vect 0$ for $j\notin\mathcal S$. The row $s/d=0$ corresponds to the null model. Results are reported for different proportions of variable types: Normal (A), Skew-Normal (B), Poisson (C), and categorical (D), under block and AR(1) correlation structures. Based on $1000$ Monte Carlo replications with sample size $n=150$.}
\label{table2}
\end{table}

In Table~\ref{table1}, we report Monte Carlo estimates of Type~I error and power under the dense contamination regime, while Table~\ref{table2} presents the corresponding results for sparse contamination. Overall, the proposed test maintains Type~I error close to the nominal level ($\alpha=0.05$) across a wide range of configurations, indicating satisfactory finite-sample calibration even when $d$ is large. The main exception is the skew-extreme setting, where empirical Type~I error exhibits moderate inflation as (d) increases. Since simulations based on the oracle precision matrix remain well calibrated in this scenario, the distortion appears to arise primarily from precision-matrix estimation rather than from the test statistic itself. A plausible explanation is that strong skewness induces heavier tails in the marginal influence scores $\hat Q_j^{(i)} $, increasing the variability of correlation-matrix estimation and making sparse precision recovery more challenging. In such settings, stronger penalization in the EBIC selection step may yield a sparser and more stable precision estimate, potentially improving finite-sample size control.

In terms of power, the test exhibits strong detection capability in both regimes. Under dense alternatives (Table~\ref{table1}), power is essentially one for moderate signal strength and remains nontrivial in weaker signal-to-noise regimes (e.g., smaller values of $k$), highlighting the ability of the procedure to accumulate weak but coordinated deviations across features. Under sparse alternatives (Table~\ref{table2}), power increases monotonically with the contamination proportion $s/d$, achieving near-perfect detection once a moderate fraction of features is affected. Importantly, even when the signal is sparse and relatively weak (small $s/d$), the test retains meaningful power, indicating sensitivity to subtle departures from the null.

Across all scenarios, performance remains stable despite substantial changes in the composition of variable types, suggesting that the proposed standardization successfully places heterogeneous features on a common scale without requiring type-specific tuning.  
Power does not increase monotonically with dimension across all contamination regimes. Under dense alternatives, its behavior depends on the rate at which the feature-level perturbations vanish: power decreases with \(d\) for \(k=2\) and \(k=3\), but remains high for \(k=4\), where the cumulative standardized signal does not vanish. Under fixed-amplitude subset contamination, by contrast, power generally increases with dimension. This behavior reflects the central mechanism of the proposed procedure: weak feature-level deviations are aggregated after standardization and dependence adjustment, allowing the cumulative signal to grow with dimension.

\begin{figure}[htp]
    \centering
\setlength{\unitlength}{1cm}
\begin{picture}(0,0)
  \put(-0.6,15.2){\rotatebox{90}{$d=100$}}
  \put(-0.6,9){\rotatebox{90}{$d=500$}}
  \put(-0.6,2.6){\rotatebox{90}{$d=1000$}}
  \put(3,19){Mean shift: $\mathcal N(3,1)$}
  \put(9.5,19){Variance shift: $\mathcal N(0,5)$}
\end{picture}
\includegraphics[width=0.9\textwidth]{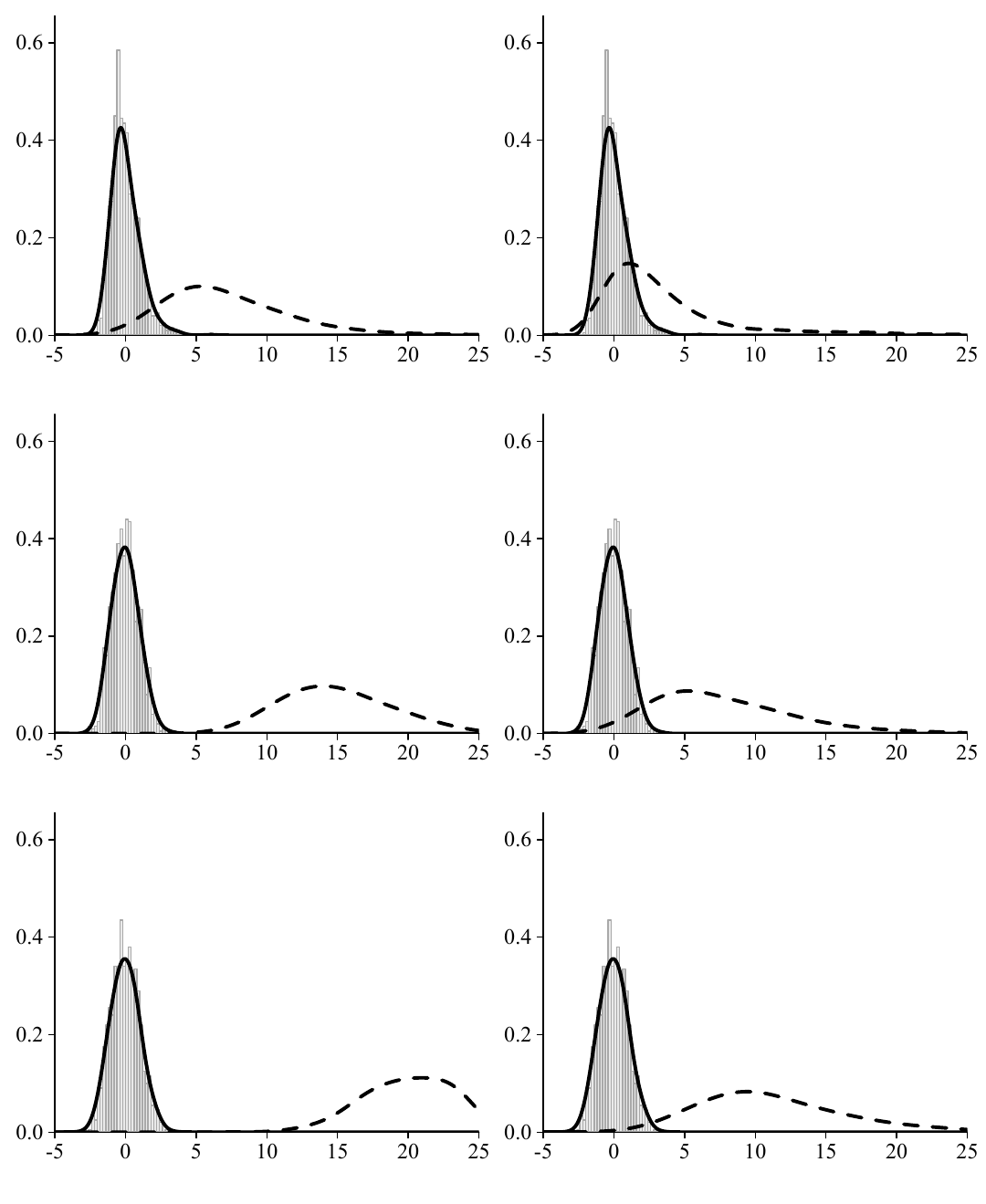}
    \caption{Densities of the standardized test statistic $MDI(\rv{\bX}^\star)$ under $H_0$ (histograms and solid kernel density estimates) and under the alternative hypothesis (dashed kernel density estimates) for $d \in \{100,500,1000\}$ (rows). The left and right columns correspond to mean and variance contamination, respectively. Contaminated features are generated from $\mathcal N(3,1)$ and $\mathcal N(0,5)$, while uncontaminated features follow $\mathcal N(0,1)$. In all settings, $5\%$ of the features are contaminated. Results are based on $1000$ Monte Carlo replications with sample size $n=750$.}
\label{fig:distributions_comparison}
    
\end{figure}

Figure~\ref{fig:distributions_comparison} illustrates the empirical null and alternative distributions of the proposed statistic under sparse contamination. Under the null hypothesis, all features are generated independently from the reference distribution \(\mathcal N(0,1)\). Under the alternative, \(95\%\) of the features remain distributed as \( \mathcal  N(0,1)\), while the remaining \(5\%\) are contaminated. In the left column, contaminated features follow \(N(3,1)\), producing a mean shift relative to the reference distribution. In the right column, contaminated features follow \( \mathcal  N(0,5)\), inducing a variance inflation while preserving the mean. Across all settings, the null distribution remains centered near zero and becomes increasingly well approximated by the standard normal distribution as \(d\) increases. At the same time, the alternative distribution shifts progressively to the right, yielding greater separation from the null distribution. This effect is evident for both contamination mechanisms and reflects the ability of the precision-weighted aggregation step to accumulate weak feature-level departures across an increasing number of variables.

\section{Real Data Examples} \label{sec:real_data}

\subsection{Anomaly Detection in Multi-omics Data} \label{sec:omics}

Multi-omics data integrate transcriptomic, epigenomic, and genomic measurements on the same individuals, providing complementary but heterogeneous information due to the coexistence of continuous, binary, and categorical variables. As an illustration, we analyze a CLL multi-omics cohort considered by \cite{lu2021}, comprising matched mRNA expression, DNA methylation measurements, and somatic mutation profiles.
IGHV mutation status is used to define two clinically meaningful comparison groups. It is a well-established prognostic marker in CLL: IGHV-mutated patients typically exhibit a more indolent disease course, whereas IGHV-unmutated patients are generally associated with more aggressive progression. Accordingly, IGHV-mutated patients form the reference group, while IGHV-unmutated patients constitute the clinically distinct evaluation group.
Each layer is reduced via marginal logistic screening with respect to IGHV status, selecting the 200 most informative RNA features, 200 methylation sites, and mutations observed in at least three patients. IGHV is excluded from the feature set, and only patients with complete measurements across all omic layers are retained. The final dataset comprises $n=184$ observations ($102$ IGHV-mutated and $82$ IGHV-unmutated) and $d=461$ mixed omic features. 

Because IGHV status is also used in the preliminary feature-screening step, this application should be interpreted as a supervised illustration of MDI rather than as an out-of-sample validation of unsupervised anomaly detection. Its purpose is to show how the method summarizes a clinically defined multivariate contrast and traces the resulting scores back to interpretable feature-level departures.

\begin{figure}[h]
\centering
\includegraphics[width=0.85\textwidth]{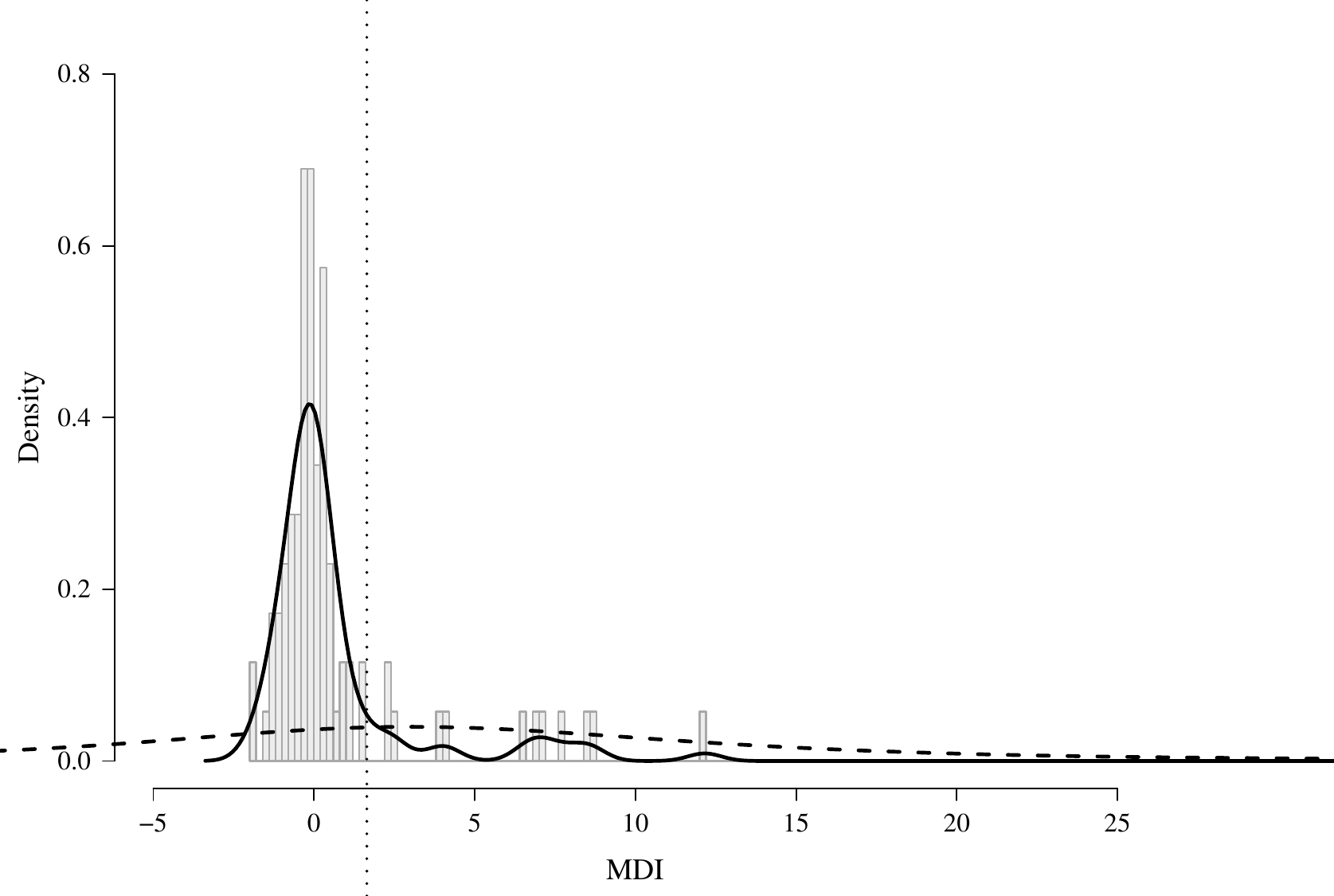}
\caption{Empirical distributions of the  MDI statistic for CLL multi-omics data subsets. Histograms and solid kernel density estimates correspond to IGHV-mutated subjects, and dashed kernel density estimates to IGHV-unmutated subjects. Scores are computed using the \(102\) IGHV-mutated subjects as the reference set.The vertical dotted line indicates the standard-normal threshold ($z_{0.90}$), corresponding to ($\alpha=0.10)$.} 
\label{fig:hist_scores}
\end{figure}

Figure~\ref{fig:hist_scores} displays the empirical distribution of the proposed MDI statistic  for the IGHV-mutated and IGHV-unmutated subjects, using the full set of 102 IGHV-mutated samples as the reference training set. For the IGHV-mutated group, the distribution is broadly centered around zero and largely consistent with the standard normal behavior expected under the null hypothesis, although some deviations remain, with a small number of subjects exhibiting relatively large scores. Such observations may reflect residual biological heterogeneity or atypical molecular profiles within the reference population. In contrast, the distribution of the statistic for the IGHV-unmutated group is clearly shifted to the right and exhibits a heavier upper tail, indicating systematically stronger departures from the reference population and providing evidence that the proposed procedure successfully distinguishes the two clinical subgroups.

We assess outlier-detection performance using IGHV mutation status as an external clinical proxy, recognizing that it captures a major but not exhaustive source of disease heterogeneity in CLL. Patients are classified as outliers when \(\operatorname{MDI}(\rv X^\star)>c_\alpha\), with \(c_\alpha=z_{1-\alpha}\), using the standard normal cutoffs corresponding to \(\alpha=0.10\) and \(\alpha=0.01\). In this setting, sensitivity, defined as the proportion of IGHV-unmutated (aggressive) patients correctly identified as outliers, provides the most relevant validation metric, while specificity and accuracy serve as complementary summaries.
Accordingly, the reported classification metrics are descriptive summaries conditional on the IGHV-guided feature selection and should not be interpreted as unbiased estimates of predictive performance.
Across \(100\) random splits, sensitivity remains high, decreasing only slightly from \(0.848\) at \(\alpha=0.10\) to \(0.822\) at \(\alpha=0.01\). As the threshold becomes more stringent, specificity increases from \(0.623\) to \(0.673\), while accuracy remains stable and slightly improves from \(0.736\) to \(0.747\). Overall, the high sensitivity indicates that a substantial fraction of detected outliers aligns with IGHV-driven disease heterogeneity, whereas the more moderate specificity is consistent with additional molecular variation not captured by IGHV status alone.

\begin{figure}[h]
\centering
\includegraphics[width=\textwidth]{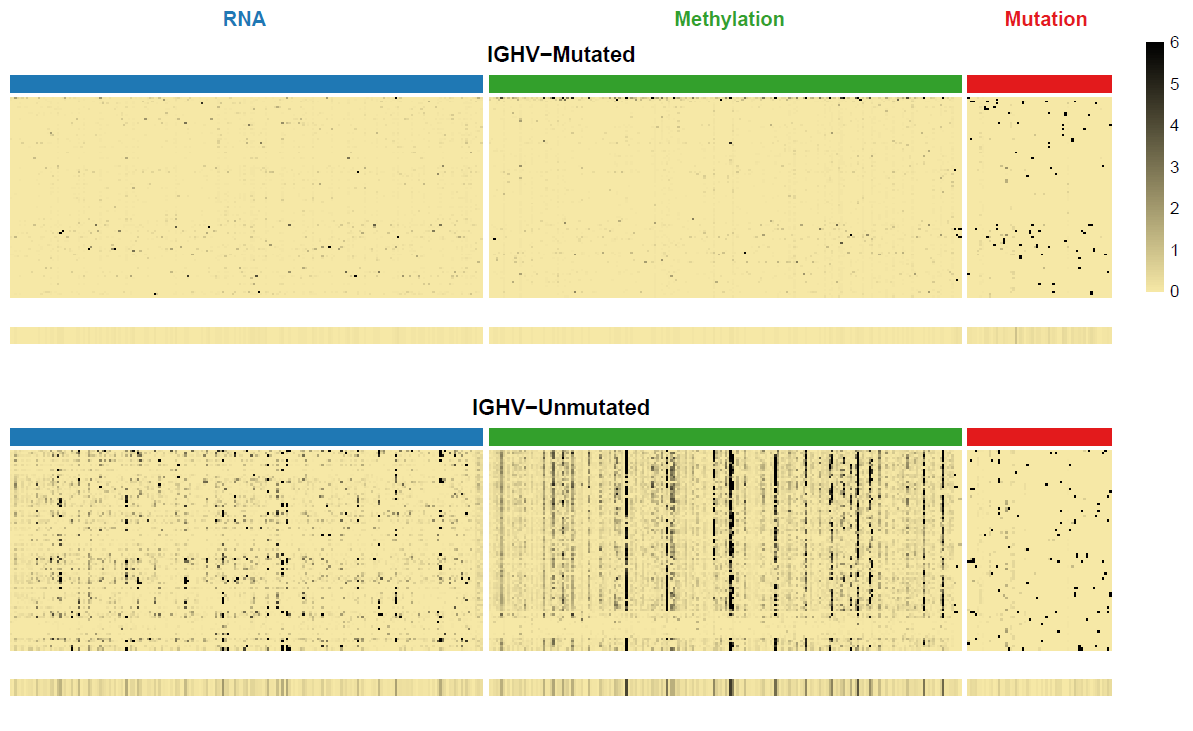}
\caption{Heatmaps for the average feature-wise contributions. For each patient   (rows) and omic feature  (columns), color intensity corresponds to the weighted absolute contribution $|\delta^{\star}_j|$ averaged across 100 random data splits. The top panel shows the IGHV-mutated group (clean data), while the bottom panel shows the IGHV-unmutated group (outliers). The bottom row of each panel reports the across-patient mean for each feature. For visualization purposes, heatmap values were capped at the 99th-percentile threshold.}
\label{fig:heatmap}
\end{figure}

To illustrate the feature-level interpretability of our procedure, we perform repeated random sample splitting on the CLL cohort. At each iteration, $20$ IGHV-mutated and $20$ IGHV-unmutated patients are randomly selected to form a balanced test set, while the remaining IGHV-mutated patients define the training (clean) reference population. The statistic is fitted on the training set and applied to all observations to compute outlier scores, with evaluation restricted to the test set. Figure~\ref{fig:heatmap} shows heatmaps of average feature-wise contributions to the test statistic across repeated sample-splitting evaluations, comparing clean (IGHV-mutated) samples and outliers (IGHV-unmutated). For each observation $i$ and feature $j$, we display the absolute weighted differences $|\delta^{\star}_j|$ averaged across 100 data splits. In the plot, columns correspond to features grouped by data type (RNA expression, DNA methylation, and mutation indicators), while rows represent patients. The weighted difference for each split is computed as $\delta^{\star}_j = \hat a_j\big(\hat Q^{\star}_j - \bar Q_j^{\mathrm{clean}}\big)$,
where $\hat Q^{\star}_j$ is the marginal quadratic influence, $\bar Q_j^{\mathrm{clean}}$ is its average over the training (clean) sample, and $\hat{\mathbf{a}} = \hat\bOmega\mathbf{1}/(\mathbf{1}^\top \hat\bOmega \mathbf{1})^{1/2}$
are weights accounting for cross-feature dependence.

The figure reveals a clear separation between clean and IGHV-unmutated samples, with stronger and more structured contributions in the latter, indicating the ability of the method to capture systematic multi-omic anomalies. The most influential features include RYR2, LAMA1, HMCN1, IRF2BP2, POT1, del18p11.3, XPO1, BIRC3, MED12, and FLRT2. These features span biologically relevant mechanisms, such as calcium signaling (RYR2), extracellular matrix organization (LAMA1, HMCN1), transcriptional and immune regulation (IRF2BP2, BIRC3), and known CLL-associated genomic alterations (POT1, del18p11.3, XPO1). Their consistent prominence across random splits indicates that IGHV-unmutated patients exhibit reproducible multi-omic deviations along these pathways, rather than isolated feature-specific anomalies.

\subsection{Anomaly Detection in ESG Data} \label{sec:esg_example}

Information on the Environmental, Social, and Governance (ESG) behavior of firms  has become an important input into investment and risk assessment decisions, with prior research linking corporate social responsibility and non-financial disclosure to access to finance, the cost of capital, and market valuation \citep{cheng2014corporate, dhaliwal2011voluntary, grewal2019market}. Anomaly detection for such data, however, remains challenging. Unlike standardized financial accounts, ESG databases combine continuous measurements, counts, binary disclosures, categorical attributes, policy indicators, and media-based controversy variables, often with substantial differences in coverage and measurement \citep{kotsantonis2019four}.

To illustrate our methodology, we use firm-level ESG data for fiscal year 2024 from LSEG Workspace\footnote{London Stock Exchange Group; \url{https://www.lseg.com/en/data-analytics/sustainable-finance/sustainability-ratings-and-data}}, accessed in April 2026. The raw data extraction contains 8,410 financial-sector firm records and 1,754 variables, including ESG indicators (e.g., emissions, workforce, and governance measures), firm-level descriptors such as country, industry, and size, and the LSEG ESG Controversies Score used to define controversy status. The latter summarizes adverse ESG-related events reported in news sources across categories such as business ethics, anti-competitive behavior, and human-rights issues, with higher values indicating lower controversy risk. Requiring an observed ESG Controversies Score leaves 1,663 firms; excluding 272 insurers yields a final sample of 1,391 non-insurance financial institutions.

ESG variables are classified as continuous, binary, or categorical and screened for redundancy using absolute Spearman correlation for continuous pairs, bias-corrected Cramér's (V) for categorical pairs, and the square root of the correlation ratio for mixed pairs. Pairs with association at least (0.85) are linked, and connected components define redundancy groups; within each group, we retain the best-covered, most granular indicator. The cutoff is prespecified and independent of controversy status and MDI results. We further exclude indicators outside the institutions' operational scope, variables based on non-comparable firm-specific scenario labels, aggregate and pillar ESG scores, and controversy-related variables. The resulting MDI input contains 404 mixed-type ESG KPIs.

Of the 1,391 firms, 1,264 have an ESG Controversies Score of 100, indicating that LSEG records no controversy, and form the candidate controversy-free reference pool; the remaining 127 firms, with scores below 100, form the evaluation group. To reduce contamination of the benchmark, we apply the MDI procedure repeatedly within the candidate reference pool using random holdouts of 50 firms. Each firm is evaluated out of sample in at least 30 holdouts and is excluded if classified as anomalous in at least half of them. This screening removes three firms, yielding a final reference sample of 1,261 institutions.

\begin{figure}[ht]
\centering
\begin{minipage}[c]{0.45\linewidth}
  \centering

  \includegraphics[width=\textwidth, trim=7mm 0mm 7mm 0mm, clip]{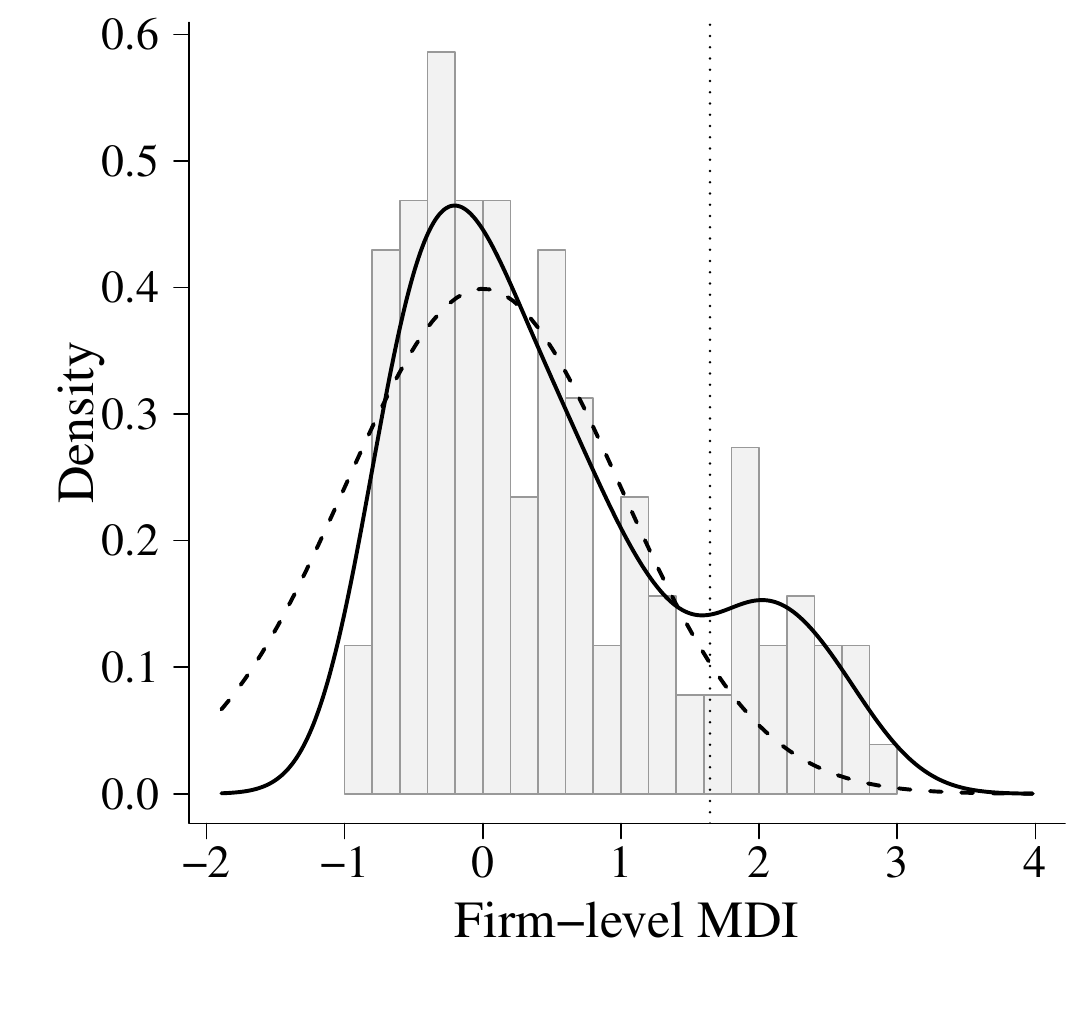}
\end{minipage}%
\begin{minipage}[c]{0.5\linewidth}
  \centering

  \raisebox{5mm}{\includegraphics[width=\textwidth,trim=20mm 0mm 12mm 0mm, clip ]{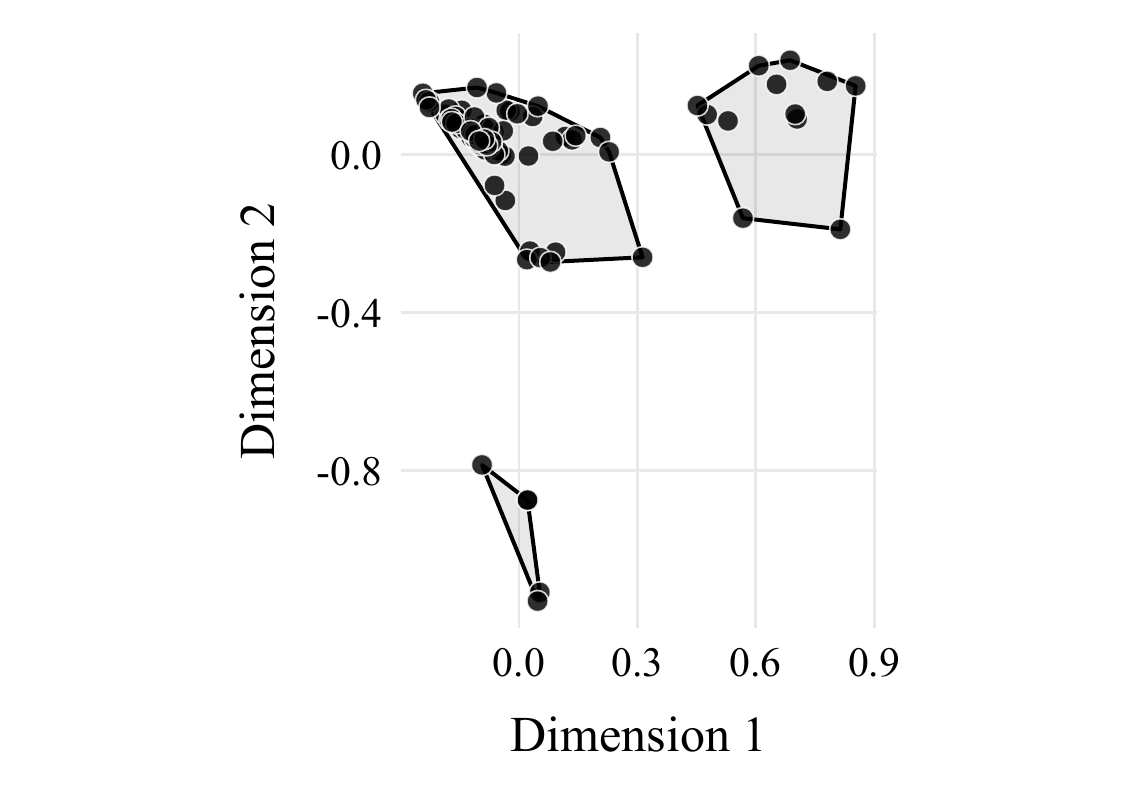}}
\end{minipage}
\caption{Analysis of the 127 controversy-exposed financial institutions. Left: distribution of firm-level MDI scores; the histogram and solid kernel density estimate correspond to controversy-exposed firms, the dashed curve denotes the standard normal density, and the vertical dotted line marks the 0.95 standard-normal quantile as a visual reference. Right: two-dimensional MDS visualization of the three clusters obtained by applying Ward's method to firms' KPI-level contribution profiles. The KPI-specific moments, covariance estimates, and dependence-adjusted aggregation weights are estimated from the controversy-free reference sample.}
\label{fig:esg_exploration}
\end{figure}

We then fit the MDI model to these 1,261 reference firms, estimating the KPI-specific moments, dependence structure, and aggregation weights, and evaluate the resulting score for all 127 controversy-exposed firms. For each firm, MDI measures the joint departure from the controversy-free reference structure and is decomposed into signed KPI-level contributions. No decision threshold is imposed in this analysis; firms are compared by the magnitude of their MDI scores. The left panel of Figure~\ref{fig:esg_exploration} shows a pronounced upper tail, indicating that a subset of controversy-exposed firms departs substantially from the reference structure while others remain comparatively close to it.

To characterize common patterns of departure, we cluster firms using their signed KPI-level MDI contributions. For firm \(i\), let $C_{ij}
=
{\hat a_j\hat Q_{ij}}/{\hat\sigma_U^{\mathrm{LOO}}}$,
and apply an \(\text{asinh}\) transformation componentwise before computing Euclidean distances. Ward's minimum-variance hierarchical method is then applied to the Euclidean distances using the \texttt{ward.D2} criterion
\citep{ward1963hierarchical,murtagh2014ward}, and the dendrogram is cut
at three clusters to obtain a parsimonious exploratory partition.
In Figure~\ref{fig:esg_exploration}, the right panel visualizes the three clusters using classical multidimensional scaling (MDS), which projects the pairwise distance matrix onto two dimensions while approximately preserving the distances between firms. Cluster membership is determined by Ward's method in the full transformed contribution space; the MDS representation is used only for visualization. The largest cluster contains 110 firms with comparatively low median MDI and diffuse contributions, while the two smaller clusters have higher MDI and distinct contribution profiles, associated respectively with workforce, climate-engagement, and human-rights indicators, and with freedom-of-expression policy, board affiliations, and capital-authorization features.

Overall, the application shows that MDI can distinguish not only the magnitude but also the structure of ESG departures from a controversy-free reference population. The global score ranks firms by the strength of their joint deviation, while the KPI-level decomposition reveals that firms with similarly elevated MDI may be anomalous for markedly different combinations of ESG characteristics. This makes the framework useful for both screening and diagnostic interpretation in ESG monitoring and risk assessment.

\section{Conclusion} \label{sec:conclusion}

The proposed MDI framework addresses the joint challenges of heterogeneous variable types, high dimensionality, cross-feature dependence, and interpretability in anomaly detection. Continuous, discrete, and categorical variables are mapped to a common scale through feature-specific moment representations, and their standardized contributions are combined through dependence-aware weighting. This allows the method to detect both concentrated departures in a few features and weaker, distributed deviations across many variables, while precision weighting limits redundancy among correlated features. The resulting global score can be used to rank observations or support threshold-based detection, while its feature-level decomposition identifies the variables and distributional characteristics driving each anomaly.

Our  numerical experiments indicate that the procedure is robust to the forms of heterogeneity that motivate the method. In particular, its performance remains stable across markedly different mixtures of continuous, discrete, and categorical variables, and across both exactly sparse and approximately sparse dependence structures. Our findings also confirm the mechanism underlying the performance of the proposed  MDI score: aggregation after feature-wise standardization allows weak marginal departures to accumulate into a detectable signal, while sensitivity to higher-order moments provides discrimination beyond classical mean-shift methods.

The two empirical applications illustrate how the method can be applied in real-world settings where mixed data types, dependence, and heterogeneous anomaly patterns arise naturally. In the multi-omics application, MDI separates IGHV-mutated and IGHV-unmutated profiles at the level of the global score while yielding feature-level contribution patterns that are biologically interpretable. In the ESG application, the method reveals substantial heterogeneity among controversy-exposed firms relative to a controversy-free reference group: some firms remain close to the reference structure, whereas others exhibit pronounced departures driven by distinct configurations of ESG indicators. Together, the applications illustrate the two complementary roles of MDI in practice: ranking observations by their overall departure from a reference population and identifying the features underlying that departure.

Several extensions are natural. Further theoretical developments could study calibration under different inferential goals and high-dimensional regimes, including the ordering of observations induced by the MDI score, as well as Gaussian, empirical, or error-controlled thresholds for binary detection. Alternative moment maps could target specific forms of anomalous behavior, while other covariance and precision-matrix estimators may be advantageous under stronger dependence or more extreme \(d/n\) ratios. These directions build naturally on the flexibility of the MDI construction, which combines heterogeneous feature representations with dependence-aware aggregation while retaining feature-level diagnostics for interpretation.

\bibliographystyle{abbrvnat}
\bibliography{biblio}

\end{document}